%% file: DAC24-ChatPattern.tex
\documentclass[sigconf,screen=true,anonymous=false,bookmarks=false,nonacm]{acmart}
\input{setting-acm}

\graphicspath{{./figs/}{../}}
\setcopyright{none}
\usepackage{mdframed}

\begin{document}

\title{
   ChatPattern: Layout Pattern Customization via Natural Language
}

\iffalse
\author{
    Zixiao Wang$^{1*}$ \quad
    Yunheng Shen$^{2*}$, \quad
    Xufeng Yao$^1$, \quad
    Wenqian Zhao$^1$, \quad
    Yang Bai$^1$, \quad
    Farzan Farnia$^1$, \quad
    Bei Yu$^1$ \\
    $^1$Chinese University of Hong Kong \quad $^2$ Tsinghua University  \\
}
\fi
\author{Zixiao Wang*}
\affiliation{\institution{CUHK}}
\author{Yunheng Shen*}
\affiliation{\institution{Tsinghua University}}
\author{Xufeng Yao}
\affiliation{\institution{CUHK}}
\author{Wenqian Zhao}
\affiliation{\institution{CUHK}}
\author{Yang Bai}
\affiliation{\institution{CUHK}}
\author{Farzan Farnia}
\affiliation{\institution{CUHK}}
\author{Bei Yu}
\affiliation{\institution{CUHK}}

\input{doc/abstract}
\maketitle
\pagestyle{empty}

\input{doc/intro}

\input{doc/prelim}

\input{doc/algo}
\input{doc/result}
\input{doc/conclusion}

% %\balance
{
\bibliographystyle{IEEEtran}
\bibliography{ref/Top-sim,ref/FPGA-DNN,ref/FPGA,ref/FPGAPlacement,ref/tools,ref/DiffPattern,ref/ChatPattern}
}

\end{document}

%% file: setting-acm.tex
\iftrue
\usepackage{titlesec}
\titlespacing\section{2pt}{3pt plus 1pt minus 1pt}{0pt plus 1pt minus 1pt}
\titlespacing\subsection{2pt}{3pt plus 1pt minus 1pt}{0pt plus 1pt minus 1pt}
\titlespacing\subsubsection{2pt}{3pt plus 1pt minus 1pt}{2pt plus 1pt minus 1pt}
\usepackage{enumitem}
\setlist{leftmargin=4.8mm}
\fi

\iftrue
\fi
\usepackage{lipsum}
\usepackage{graphicx}
\usepackage{amsmath}
\usepackage{footnote}
\usepackage{mathtools}
\usepackage{mathrsfs}     % mathscr command
\usepackage{comment}
\usepackage[subrefformat=parens,labelformat=parens]{subfig}
\usepackage{bm,bbm}
\usepackage{multirow}
\usepackage{threeparttable,booktabs}
\usepackage{blkarray}
\usepackage{tikz}
\usetikzlibrary{positioning,calc,fit,decorations.pathmorphing,shapes.geometric,shapes.gates.logic.US,calc}
\usepackage{tikz-3dplot}
\usepackage{balance}
\usepackage{courier}                                       % courier font, used in \texttt
\usepackage{cleveref}                                      % smart citation
\usepackage[mathcal]{eucal}
\usepackage[]{algpseudocode}                               % algorithm package
\algrenewcommand\textproc{\texttt}
\makeatletter\let\float@addtolists\relax\makeatother
\usepackage{algorithm}
\usepackage{filecontents}                                  % support to pgfplots
\usepackage{pgfplots}
\usepackage{pgfplotstable}
\usepackage{pgf-pie}
\usepgfplotslibrary{groupplots}
\pgfplotsset{compat=newest}
\usepackage[figuresright]{rotating}
\usepackage{xcolor,colortbl}                               % \rowcolor
\newcommand{\tool}[1]{$\mathsf{#1}$}

\renewcommand{\vec}[1]{\boldsymbol{#1}}

\newcommand{\minisection}[1]{\vspace{.06in}\noindent{\textbf{#1}}}

\theoremstyle{plain}

\theoremstyle{definition}
\newtheorem{mydefinition}{\textbf{Definition}}
\newtheorem{myproblem}{\textbf{Problem}}

\algrenewcommand\textproc{\texttt}

\usepackage[skip=1pt]{caption}            % set space between figure and caption
\definecolor{CUHKorange}{RGB}{244,106,18} %F47012
\definecolor{CUHKblue}{RGB}{0,111,190}    %006FBE
\definecolor{CUHKgreen}{RGB}{0,127,128}   %007F80
\definecolor{CUHKred}{RGB}{228,46,36}     %E42E24
\definecolor{CUHKyellow}{RGB}{198,148,34} %C69422
\definecolor{CUHKdark}{RGB}{114,44,114}   %722C72
\definecolor{CUHKmiddle}{RGB}{144,44,144} %902C90

\usepackage{tcolorbox}
\tcbuselibrary{skins,breakable}
    {\endtcolorbox}

%% file: doc/abstract.tex
\begin{abstract}
Existing works focus on fixed-size layout pattern generation, while the more practical free-size pattern generation receives limited attention. In this paper, we propose ChatPattern, a novel Large-Language-Model (LLM) powered framework for flexible pattern customization. ChatPattern utilizes a two-part system featuring an expert LLM agent and a highly controllable layout pattern generator. The LLM agent can interpret natural language requirements and operate design tools to meet specified needs, while the generator excels in conditional layout generation, pattern modification, and memory-friendly patterns extension. Experiments on challenging pattern generation setting shows the ability of ChatPattern to synthesize high-quality large-scale patterns.

\end{abstract}
    

%% file: doc/intro.tex
\section{Introduction}
\label{sec:intro}

High-quality Very-Large-Scale Integration (VLSI) layout pattern libraries are foundational to numerous Design for Manufacturability (DFM) studies, such as refining design rules, formulating Optical Proximity Correction (OPC) recipes \cite{gao2014mosaic}, conducting lithography simulations \cite{kuang2013efficient,yu2015layout}, and detecting layout hotspots \cite{chen2019faster}. Contemporary machine-learning-based lithography design applications typically require an extensive array of layout patterns to train their networks, yet assembling a large-scale pattern library is cost-prohibitive due to the intricate logic-to-chip design cycle.

Prior to the advent of machine learning, rule-based methods \cite{reddy2018enhanced} were employed to synthesize layout patterns automatically, with simple hand-drafted augmentations like flipping and rotation used to expand the fundamental layout pattern units. These units would then be assembled randomly to create larger designs. However, the diversity and volume of patterns generated by these rule-based methods often fell short of user expectations. 
Conversely, recent learning-based methods \cite{yang2019deepattern, zhang2020layout, wen2022layoutransformer, wang2023diffpattern} have demonstrated the ability to generate a plethora of diverse layout patterns that closely match the dataset distribution. One notable technique, the squish pattern trick \cite{gennari2014topology}, has been instrumental in reducing computational and memory demands by condensing a layout pattern patch into a compact 2D-topology matrix, upon which generative models are trained.
Once the topology matrices are generated, legalization technologies are employed to recover the topology matrices to layout patterns.  

Despite these advancements, learning-based methods still lack the capability for fine-grained modifications. For instance, when a training set comprises various pattern categories, these methods cannot determine the class of the generated pattern. Editing a specific section of a layout according to particular rules is a frequent necessity, especially with large layouts, to accommodate specific applications \cite{sun2022efficient,zhao2022adaopc}, but none of existing methods support pattern edition. And the size of generated topology matrix is usually fixed due to the limitation of network architecture and device memory. The fixed pattern size restricts the application of the generated patterns in downstream tasks which requires different pattern size. 

Additionally, the task of manually synthesizing millions of desired layout patterns using flexible generation tools can be labor-intensive. Pattern library builders must not only master complex generation tools but also comprehend the specialized requirements from downstream users, often communicated in a blend of natural language and professional jargon. Moreover, builders may not always be available or able to respond promptly. To bridge this gap, Large Language Models (LLMs) have proven adept at handling complex tasks across various domains \cite{zhou2023dbgpt,he2023chateda}. However, their application as layout pattern library builders remains unexplored.

\begin{figure}[tb!]
    \centering
    \includegraphics[width=0.85\linewidth]{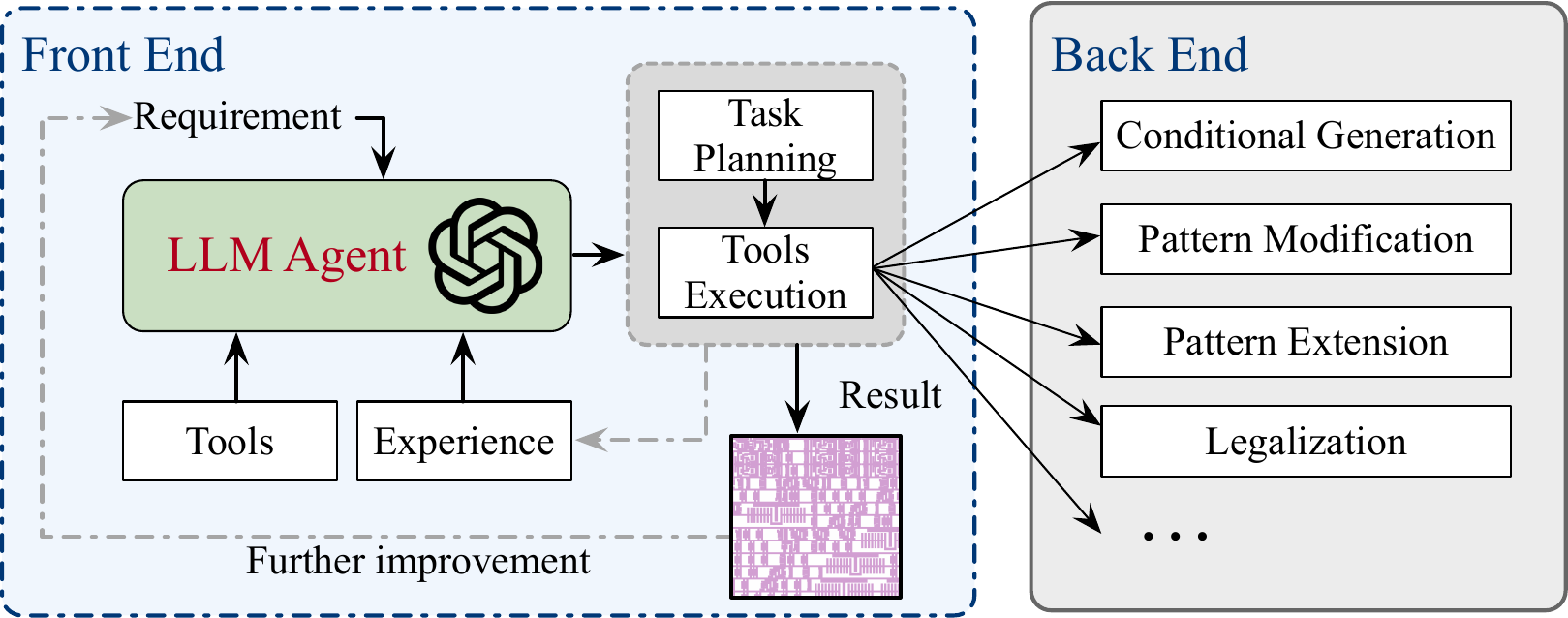}
    \caption{An illustration of ChatPattern.}
    \label{fig:chatpattern}
\end{figure}

In response, this paper introduces \tool{ChatPattern}, an indefatigable layout pattern builder designed to tailor a pattern library to specific requirements articulated in natural language. As illustrated in \Cref{fig:chatpattern}, \tool{ChatPattern} is composed of two principal components: an expert LLM agent holding design tools and documentation, capable of understanding and executing tasks based on natural language instructions, and a flexible, controllable layout pattern generative model that surpasses existing methods by offering conditional layout generation, precise pattern modification, and unrestricted pattern extension.
The contributions of this work are fourfold:
\begin{itemize}
    \item The presentation of ChatPattern, the inaugural LLM-powered layout pattern generation framework.
    \item The integration of an expert LLM agent as a pattern library builder, proficient in processing natural language inputs and autonomously operating the necessary tools to fulfill requirements.
    \item The development of a versatile layout pattern generative model that outperforms existing methods in conditional pattern generation, layout modification, and free-size pattern extension.
    \item The expansion of the scope of the layout pattern generation task, prompting researchers to focus on more realistic yet challenging tasks such as free-size layout pattern generation.
\end{itemize}

%% file: doc/prelim.tex
\section{Preliminaries}
\subsection{The Scope of \tool{ChatPattern}}
\tool{ChatPattern} is an AI agent that offers a conversational interface, enabling users to use natural language to guide the creation of pattern libraries that meet their specific layout generation needs.

\begin{myproblem}[Target of \tool{ChatPattern}]
    Create a framework powered by a large language model that can understand and process natural language instructions, and produce a library of legal layout patterns that meet user needs by utilizing a layout pattern generation tool.
\end{myproblem}

\subsection{Fixed-size Layout Pattern Generation}

Contemporary research in layout pattern synthesis predominantly focuses on creating patterns of a predetermined dimension. This research domain has witnessed significant contributions that leverage squish-pattern representation \cite{yang2019deepattern,zhang2020layout,wang2023diffpattern}. These methods learn to fit the distribution of the binary topology matrix of Squish Pattern~\cite{gennari2014topology}. And the generated topology matrix will be further legalized in post-processing to yield a legal layout pattern.

\begin{figure}[tb!]
    \centering
    \includegraphics[width=0.76\linewidth]{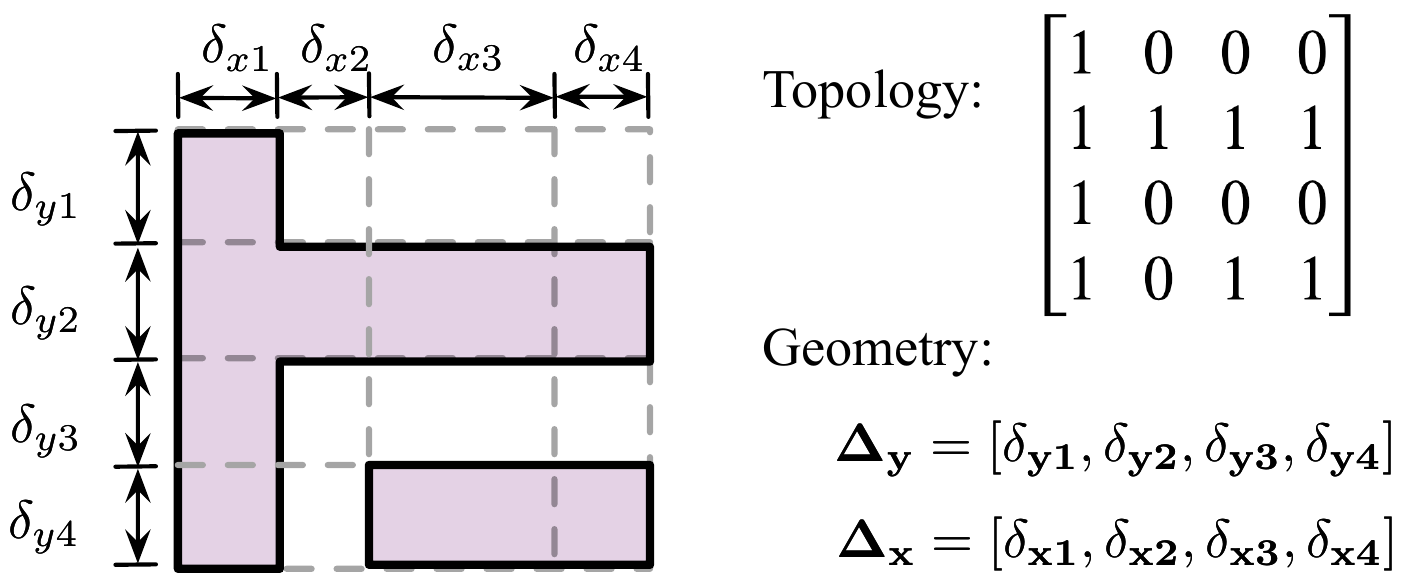}
    \caption{Squish Pattern Representation.}
    \label{fig:squishpattern}
\end{figure}

\minisection{Squish Pattern Representation}.
A layout pattern, comprising a series of non-overlapping polygons, can be effectively modeled using a squish pattern. The squish pattern is a compact representation of layout pattern that encodes layout topology and geometry into a matrix $\vec{T}$ and vectors $\bm \Delta_x,\, \bm \Delta_y$. It divides the layout into grids using scan lines along polygon edges, storing intervals in the $\bm \Delta$ vectors. Matrix entries are binary, denoting shapes or emptiness. Topology matrix will be further normalized to a fixed-size square for uniformity as introduced in \cite{yang2019detecting}.
\Cref{fig:squishpattern} illustrates how squish pattern works.

\minisection{Topology Generation via Unconditional Discrete Diffusion.}
Discrete diffusion model~\cite{austin2021structured} is a type of generative model where the value of every image pixel is limited in a pre-defined discrete space.
Similar to normal diffusion models~\cite{ho2020denoising,song2020denoising}, noise is added to input images in forward process, and a model with learnable parameter $\theta$ learns to remove noise in reverse process.
Given that $x_k\in \{0,1\}$ is an entry in the topology matrix $\vec{T}$, a transition probability matrix $[\vec{Q}_k]_{ij} = q(x_k=j|x_{k-1}=i)$ is defined to describe the state transition probability for each $\vec{x}$ at the $k$-th of $K$ forward step,
\begin{equation}
\vec{q}\left(\vec{x}_k \mid \vec{x}_{k-1}\right) := \operatorname{Cat}\left(\vec{x}_k ; \vec{p}=\vec{x}_{k-1} \vec{Q}_k\right),
\label{eq:d3pm-forward}
\end{equation}
where $\operatorname{Cat(\vec{x}|\vec{p})}$ denote the categorical distribution of $\vec{x}$ given $\vec{p}$. And the forward process can be defined as,
\begin{align}
    \vec{q}\left(\vec{x}_k|\vec{x}_0\right) &= \operatorname{Cat}\left(\vec{x}_k ; \vec{p}=\vec{x}_0 \overline{\vec{Q}}_k\right), \label{eq:forward} \\
    \vec{Q}_k &= \begin{bmatrix} 1-\beta_k & \beta_k \\ \beta_k & 1-\beta_k \end{bmatrix}, \\
    \beta_k &= \frac{(k-1)\left(\beta_K-\beta_1\right)}{K-1}+\beta_1,~k = 1,...,K,
\end{align}
where $\overline{\vec{Q}}_k=\vec{Q}_1 \vec{Q}_2 \ldots \vec{Q}_k$ and $\beta_1$ and $\beta_K$ are hyper-parameters.

After sampling a noised topology $\vec{T}_k$ from the input topology $\vec{T}_0$ via \Cref{eq:forward}, the model learns to denoise the topology and predicts the logits of the posterior distribution $\vec{p}_{\vec{\theta}}\left(\vec{x}_0|\vec{x}_k\right)$. Therefore, the $k$-th reverse step can be calculated as following:
\begin{equation}
\vec{p}_{\vec{\theta}}\left(\vec{x}_{k-1} | \vec{x}_k\right) = \sum_{\widetilde{\vec{x}}_0} \vec{q}\left(\vec{x}_{k-1} | \vec{x}_k , \widetilde{\vec{x}}_0\right) \vec{p}_{\vec{\theta}}\left(\widetilde{\vec{x}}_0 | \vec{x}_k\right),
\label{eq:d3pm-reverse}
\end{equation}
where the term $\widetilde{\vec{x}}_0$ will visit every possible state of $\vec{x}_0$. 
% The training process will optimize a $\lambda$-weighted combination of a variational upper bound on the negative log-likelihood and the prediction error,
% \begin{equation}
% L = D_{\mathrm{KL}}\left(\vec{q}\left(\vec{x}_{k-1} | \vec{x}_{k}, \vec{x}_0\right) \parallel \vec{p}_\theta\left(\vec{x}_{k-1} | \vec{x}_{k}\right)\right) - \lambda\log \vec{p}_{\vec{\theta}} \left(\vec{x}_0 | \vec{x}_k\right),
% \label{eq:d3pm-loss-true}
% \end{equation}
% where $\vec{q}\left(\vec{x}_{k-1} | \vec{x}_k, \vec{x}_0\right)$ has a closed form according to \Cref{eq:d3pm-forward} and Bayes' theorem. 
Finally, we can denoise a sampled noise $\vec{T}_K$ and synthesize a topology $\vec{T}_0$ with the well-trained model by recursively calling the reverse step,
\begin{equation}
    p_\theta(\vec{T}_0|\vec{T}_K) =p_\theta(\vec{T}_{0}|\vec{T}_1) \prod_{k=2}^{K}p_\theta(\vec{T}_{k-1}|\vec{T}_k).
\end{equation}

\begin{figure}[tb!]
    \centering
    \includegraphics[width=0.78\linewidth]{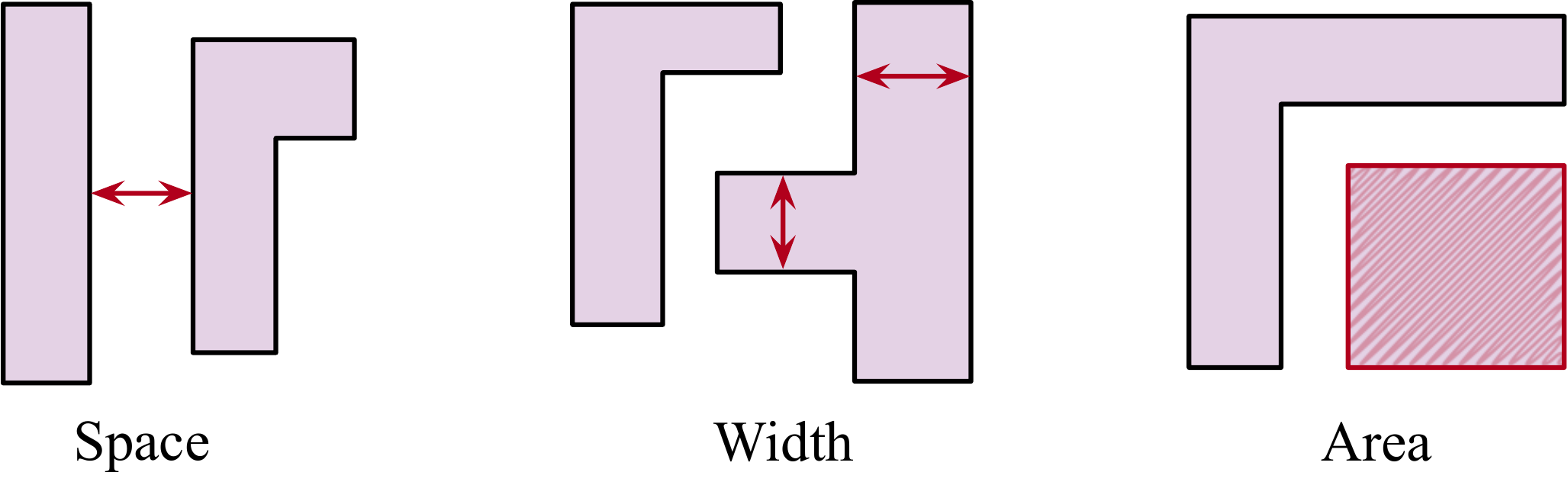}
    \caption{
        Design rule illustrations.
        `Space' means distance between adjacent polygons.
        `Width' measures shape size in one direction.
        `Area' denotes area of a polygon. }
    \label{fig:drc_rule}
\end{figure}

\minisection{Topology Legalization}. The generated topology matrix will be further enriched to patterns by matching them with suitable geometry vectors.
The generated patterns should satisfy several pre-defined design rules of IC layout~\cite{zhang2020layout,wen2022layoutransformer} as illustrated in \Cref{fig:drc_rule}. 

\begin{mydefinition}[Legality]
We treat a layout pattern as a legal one if the layout pattern is DRC-clean, given the design rules.
\end{mydefinition}
\begin{equation}
    \text{Legality} = \frac{\# \text{Legal Patterns}}{\# \text{Generated Patterns}}
    \label{eq:legality}
\end{equation}

\begin{figure*}[tb!]
    \centering
    \includegraphics[width=0.88\linewidth]{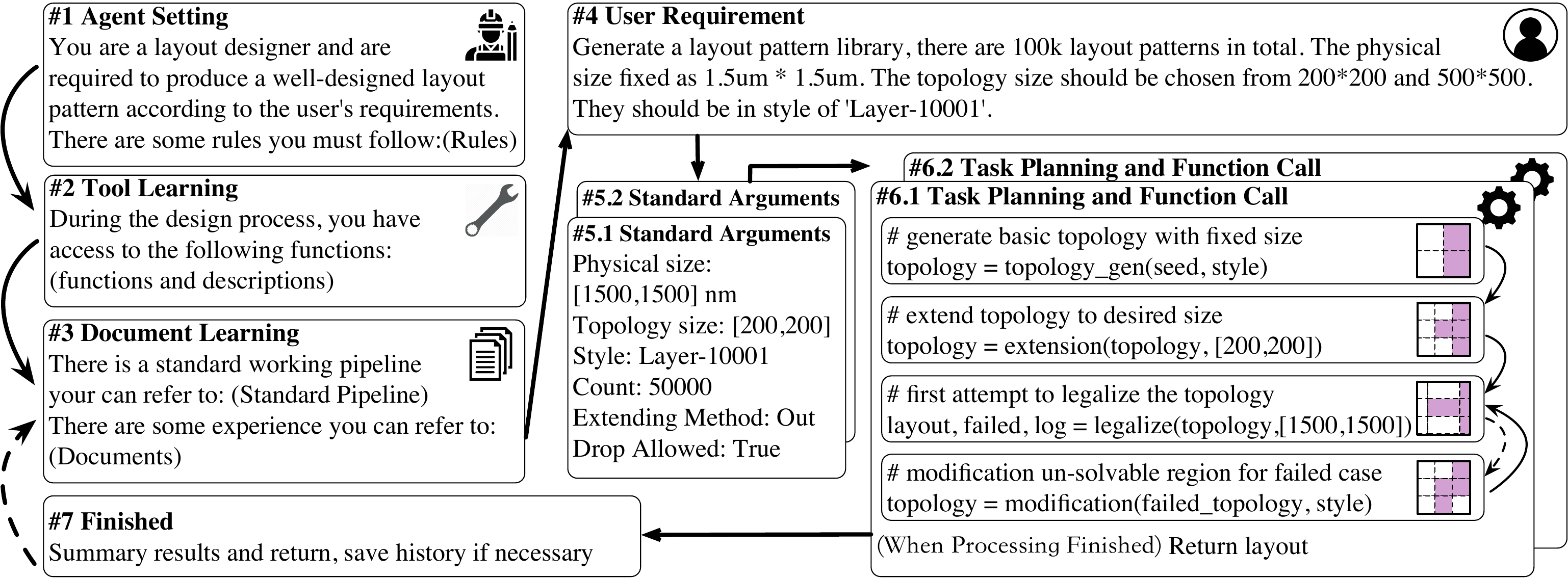}
    \caption{A working pipeline of LLM agent in \tool{ChatPattern}. Dash lines denote the optional paths.}
    \label{fig:llm}
\end{figure*}

\subsection{Free-size Layout Pattern Generation}
Compared with fixed-size layout pattern generation task, we aim to synthesize layout patterns with any given size, which is a more challenging task given the limited model output size, device memory, and varied requirements. We adopt the widely-used evaluation metric~\cite{yang2019deepattern,wang2023diffpattern,wen2022layoutransformer,zhang2020layout}, Diversity, to evaluate the quality of generated pattern library.
A greater pattern diversity $H$ indicates that the library contains more widely distributed patterns.
\begin{mydefinition}[Diversity]
The diversity of the patterns library, denoted by $H$, is defined as the Shannon Entropy of the distribution of the pattern complexities as follows:
\begin{equation}
    H = - \sum_i \sum_j P(c_{xi}, c_{yj})\log{P(c_{xi}, c_{yj})},
    \label{eq:diveristy}
\end{equation}
where $P(c_{xi}, c_{yj})$ is the probability of a pattern with complexity $(c_{xi}, c_{yj})$ sampled from the library, and $c_x$ and $c_y$ are the numbers of scan lines subtracted by one along the x-axis and y-axis, respectively.
\end{mydefinition}

Based on the above evaluation metrics, the pattern generation problem can be formulated as follows,

\begin{myproblem}[Free-size Layout Pattern Extension]
    Given a set of design rules and target pattern size, the objective of pattern generation is to synthesize a legal pattern library such that the pattern diversity of the layout patterns in the library is maximized.
\end{myproblem}

%% file: doc/algo.tex
\section{Algorithm}
\label{Sec:algo}

%\subsection{Overview of \tool{ChatPattern}}
\tool{ChatPattern} seamlessly integrates a front-end powered by a Large Language Model with a back-end that employs a conditional discrete-diffusion model for layout pattern generation.
% This LLM agent excels at natural language communication, adeptly understanding user requirements, and orchestrating scripts to efficiently generate a pattern library.
% The pattern generative model, providing API functions for LLM agent, is specifically designed for tasks involving free-size pattern generation.

\subsection{Pattern Customization via Expert LLM agent}
\tool{ChatPattern}, leveraging LLM technology, automates pattern library customization. Its primary function is to interpret user requirements expressed in natural language, breaking down complex demands into manageable sub-tasks. These sub-tasks are subsequently processed using specialized pattern generation tools. The detailed operational pipeline of \tool{ChatPattern} is illustrated in \Cref{fig:llm}. 

\minisection{Requirement Auto-Formatting.}
Given the complexity of user requests in the Layout Pattern Generation workflow, the LLM agent first translates these requests into a structured format using a pre-defined template. This template contains all relevant parameters and requirements for utilization by the pattern generation API functions. The template can be summarized by the LLM agent according to the description of functions or provided by the function provider. The arguments can be further divided into required ones, which decide the basic parameter (\textit{e.g.,} pattern count and size) of this task, and optional ones, which are for fine-grained control. An important thing is that one task given by a user can be decomposed into several requirement lists, each for one simple sub-task, to limit the complexity of one task and reduce mistakes. For example, the user request shown in \Cref{fig:llm} is factorized into two sub-tasks.

\minisection{Task Planning and Execution.}
During the last step, \tool{ChatPattern} identifies and plans the necessary sub-tasks for fulfilling the user's request. For each planned sub-task, \tool{ChatPattern} schedules a series of structured tasks, which are then addressed using various pattern generation tools like topology generation and legalization. The task planning process is exemplified in \Cref{fig:llm}. For execution, \tool{ChatPattern} utilizes a Python interpreter or API calls, supported by the layout pattern generation back-end model.
Compared with a rule-based arguments-program translator, the LLM agent can address failed cases based on feedback or logs derived from function calls. For example, in the legalization phase, the legalization can easily fail especially for large topology matrices due to the violation of design rules. Applying topology selection and dropping the failed cases can guarantee the legality of the final result, while a lot of effort on topology matrix generation is wasted. Alternative choices for LLM agent are modifying the failed regions with different conditions and trying different initial states, which save time and computation.

\minisection{Tool Function Learning and Application.}
The core concept of pattern generation through an LLM agent lies in its ability to operate without directly accessing the generated matrix of zeros and ones. This matrix could surpass the token length limit of the LLM, rendering it incapable of extracting information from such low-level data. The LLM agent instead relies on the route to the end results and comprehension of select overarching characteristics, such as complexity, physical dimensions, or error locations. To construct a pattern library, certain fundamental tools or APIs are indispensable: (1) \noindent\textit{Random Topology Generation}, which creates random topologies subject to specific conditions, and (2) \noindent\textit{Topology Legalization}, which transforms a topology into a compliant layout pattern. These functions constitute the cornerstone for squish-pattern-based layout pattern generation. Yet, the scope of the generated topology might be constrained by the model size or device capacity, and it might not always conform to the legalization process. Thus, supplementary functions for advanced customization are available: (1) \noindent\textit{Topology Extension} augments a topology to a designated size within certain parameters, and (2) \noindent\textit{Topology Modification} revises parts of a topology, offering a time-efficient alternative to discarding non-compliant topologies, particularly for expansive patterns. 
Additional high-level functions may be integrated to bolster the model’s capabilities, allowing the LLM agent to capitalize on backend tool enhancements with no need for internal changes.

\minisection{Learning from Documents and Experience.} Documents are reservoirs of high-level knowledge, instrumental in the pattern design process. In the free-size layout pattern generation task, the selection of the extension algorithm plays a pivotal role in determining the quality of the output. Informed algorithm selection can be enhanced through statistical analysis of historical data. Additionally, equipping a model with a standardized operational pipeline at inception can lead to a marked increase in performance. The practice of documenting work history and scrutinizing exceptional cases is equally valuable, laying the groundwork for the model's ongoing refinement.
% The documents contain high-level knowledge which can be useful during the pattern design. For example, in the free-size layout pattern generation task, the quality of the generated layout pattern can be partly decided by the choice of the extension algorithm. Statistical analysis of previous work history can facilitate better algorithm selection. Another example is that, for a model's initial operation, providing a standard working pipeline can significantly improve performance. Recording the working history and analyzing rare cases also contribute to the model's continuous improvement.

\subsection{Flexible Layout Pattern Generative Model}
\label{sec:3.3}

The pattern generator provides tool support for LLM agent front-end. Our generative model utilizes a conditional discrete diffusion model to fit the topology distributions from multi-source and synthesize topology for further legalization. However, compared with previous work, \tool{ChatPattern} leverages three novel characters to enable the ability to generate complex topology matrix and meet highly customized requirements.

\begin{figure}[tb!]
    \centering
    \includegraphics[width=0.88\linewidth]{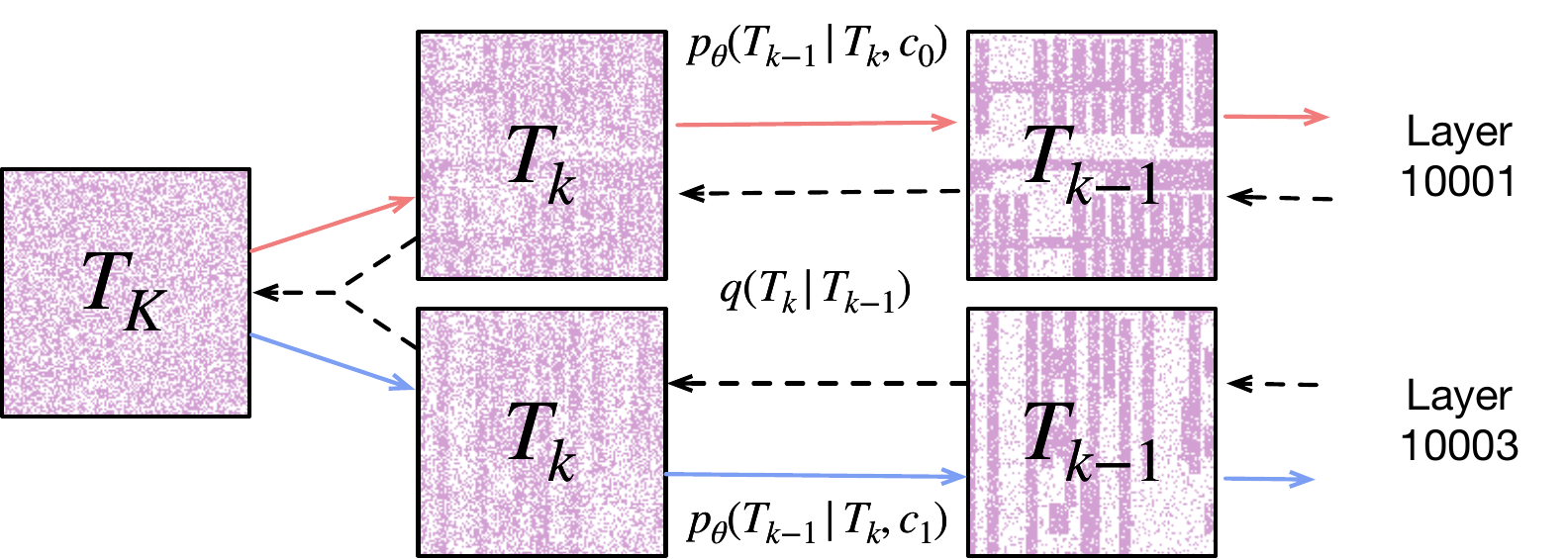}
    \caption{Illustration of conditional pattern generation.}
    \label{fig:generation}
\end{figure}

\minisection{Property-Conditional Topology Generation.} Patterns in a dataset do not always share the design rules or physical properties. For example, the distribution of patterns from the 7-nm manufacturing process is different from that of 130-nm. The diversity of materials and design rules etc. enlarges the difference. Establishing a dataset and training individual models for each kind of patterns is not a sound choice, while directly training a model on the mixture dataset will raise some concerns about the design rules conflict. To address this issue, we add conditions to the reverse process of the diffusion model and the distribution of the generated topology matrix can be specified. Different from the cases in normal image generation, the condition design in pattern generation should consider the design rules, materials, and manufacturing process. The $k$-th step of distribution specified reverse process can be defined as,
\begin{equation}
\vec{p}_{\vec{\theta}}\left(\vec{x}_{k-1} | \vec{x}_k,c\right) = \sum_{\widetilde{\vec{x}}_0} \vec{q}\left(\vec{x}_{k-1} | \vec{x}_k , \widetilde{\vec{x}}_0\right) \vec{p}_{\vec{\theta}}\left(\widetilde{\vec{x}}_0 | \vec{x}_k,\vec{c}\right),
\end{equation}
where $\vec{c}$ denotes the conditions. An illustration of the training and sampling process is shown in \Cref{fig:generation}. During the training process, the model optimizes the loss function,
\begin{equation}
    L = D_{\mathrm{KL}}\left(\vec{q}\left(\vec{x}_{k-1} | \vec{x}_{k}, \vec{x}_0\right) \parallel \vec{p}_\theta\left(\vec{x}_{k-1} | \vec{x}_{k},\vec{c}\right)\right) - \lambda\log \vec{p}_{\vec{\theta}} \left(\vec{x}_0 | \vec{x}_k,\vec{c}\right).
\end{equation}
When the training process is finished, a topology matrix with condition $\vec{c}$ can be generated by a $K$-step reverse process,
\begin{equation}
    p_\theta(\vec{T}_0|\vec{T}_K,\vec{c}) =p_\theta(\vec{T}_{0}|\vec{T}_1,\vec{c}) \prod_{k=2}^{K}p_\theta(\vec{T}_{k-1}|\vec{T}_k,\vec{c}),
    \label{eq:c-rev}
\end{equation}
where $\vec{T}_K$ is a randomly-sampled noise.

\begin{figure}[tb!]
    \centering
    \includegraphics[width=0.88\linewidth]{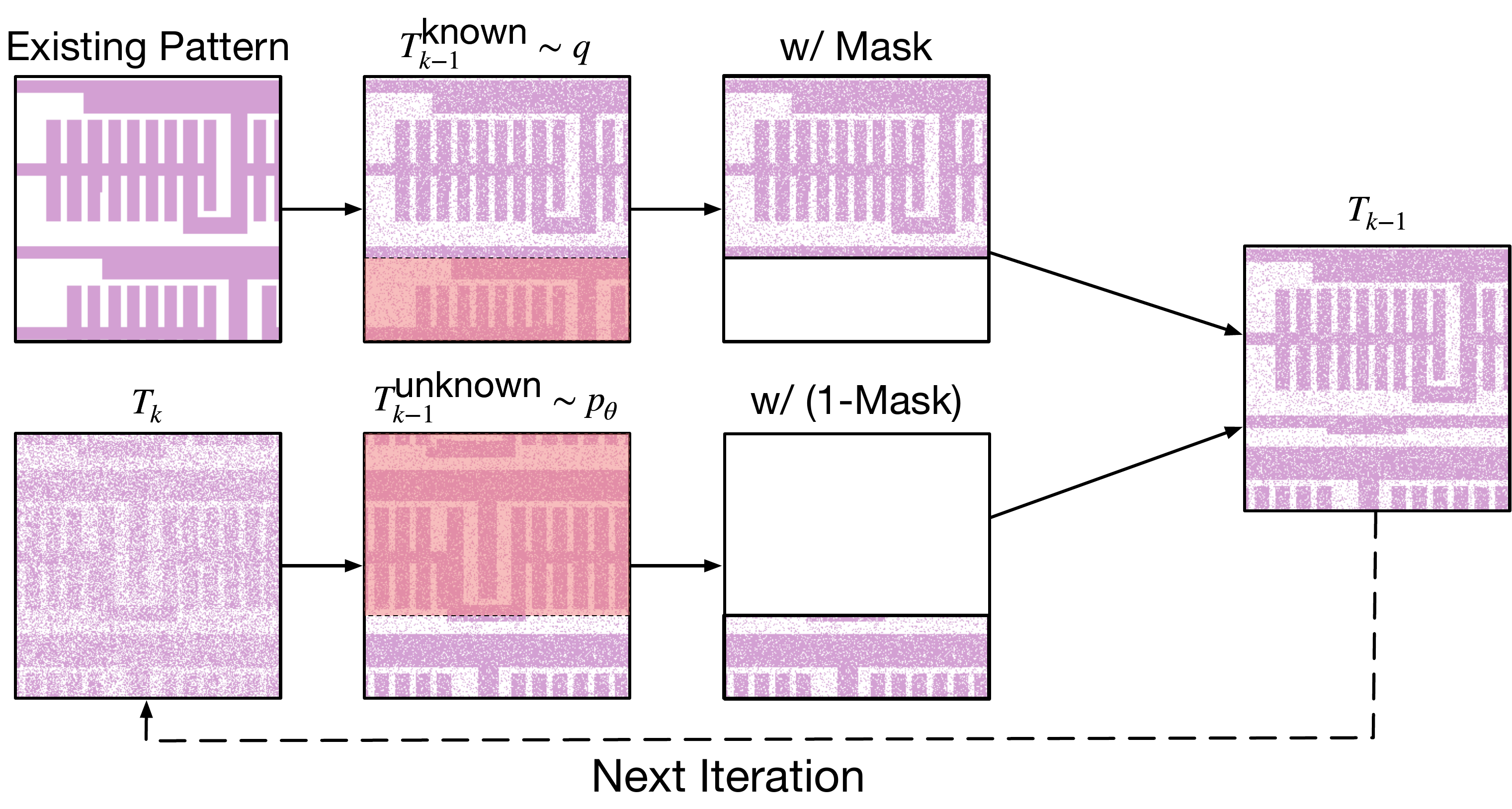}
    \caption{Illustration of pattern modification.}
    \label{fig:inpainting}
\end{figure}

\minisection{Pattern Modification.}
Given an existing pattern topology matrix $\vec{T}_0^{\text{known}}$, making modifications to any desired region on it can be useful when dealing with failed topology.
We denote kept pixels as $\vec{M}\odot\vec{T}_0^{\text{known}}$ and denote the masked pixels as $(1-\vec{M})\odot\vec{T}_0^{\text{known}}$. Since in every reverse step of \Cref{eq:c-rev}, $\vec{T}_{k-1}$ depends solely on $\vec{T}_k$, we replace the kept region in $\vec{T}_k$ by the noised pixels in given topology, $\vec{M}\odot\vec{T}_k^{\text{known}}$. The noised topology $\vec{T}_k^{\text{known}}$ is obtained by the forward process in \Cref{eq:forward}. And $(1-\vec{M})\odot\vec{T}_{k-1}^\text{unknown}$ will be generated by the model with the condition of $\vec{M}\odot\vec{T}_k^{\text{known}}$. And different with general image generation\cite{ho2020denoising,lugmayr2022repaint}, the modification of patterns should consider the design rules and the condition $\vec{c}$ should match the distribution of given topology matrix $\vec{T}_0^{\text{known}}$,
\begin{equation}
    \begin{aligned}
    \vec{T}_{k-1}^\text{known} &\sim \vec{q}\left(\vec{T}_{k-1}|\vec{T}_0^\text{known}\right), \\
    \vec{T}_{k-1}^\text{unknown} &\sim \vec{p}_{\vec{\theta}}\left(\vec{T}_{k-1} | \vec{T}_k,\vec{c}\right),  \\
    \vec{T}_{k-1} &= \vec{M} \odot \vec{T}_{k-1}^\text{known} + (1-\vec{M}) \odot \vec{T}_{k-1}^\text{unknown},
\end{aligned}
\end{equation}
where $\vec{T}_0^\text{known}$ shares the design rules with patterns in condition $\vec{c}$. An illustration of pattern modification can be found in \Cref{fig:inpainting}.

\begin{figure}[tb!]
    \centering
    \includegraphics[width=0.80\linewidth]{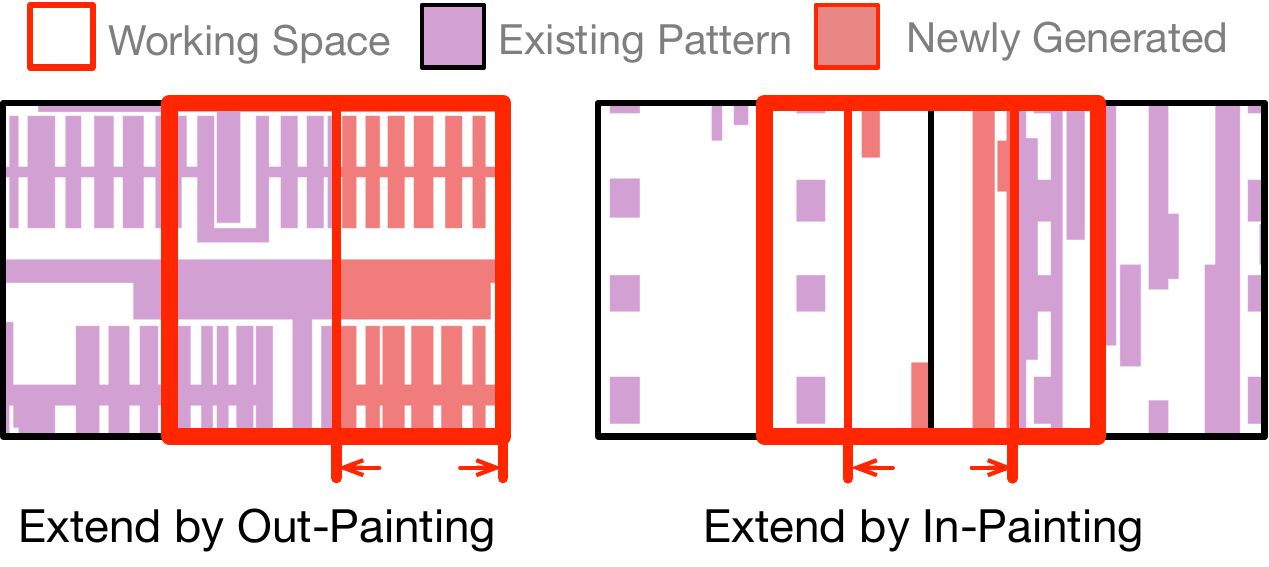}
    \caption{Pattern extension via In-Painting and Out-Painting.}
    \label{fig:outpainting}
\end{figure}

\minisection{Pattern Extension.}
Extending a given pattern to a larger one is a practical function since the model output usually takes a fixed size while the required patterns can vary among a large range.
Pattern extension can be achieved via In-Painting and Out-Painting with pattern modification techniques. An illustration of In-Painting extension and Out-Painting extension can be found in \Cref{fig:outpainting}.
By modifying the adjacency border and corner of a concatenated topology matrix, we can merge the shape from both sides and synthesize a larger topology matrix, which we denoted as In-Painting. 
We treat the extending method that directly generates a new border of an existing pattern as Out-Painting. 

By recursively extending a given pattern, we can extend the pattern to any desired size without considering memory limitation, since only the region in working space will be taken into computation. 
If we assume the target size of topology is $[W,H]$ and the window size of the model is $[L,L]$, the number of sampling for In-Painting can be calculated by, $N_{\text{in}}=(2\lceil\frac{W}{L}\rceil-1)(2\lceil\frac{H}{L}\rceil -1)$.
With a stride $S$, the number of sampling for Out-Painting is $N_{out}=(\lceil\frac{W-L}{S}\rceil + 1)(\lceil\frac{H-L}{S}\rceil + 1)$.

\minisection{Legalization.} We utilize the non-linear legalization function proposed in DiffPattern\cite{wang2023diffpattern} and denoted it as,
\begin{equation}
    \text{Legalization}(\cdot) = f_{\mathcal{R}}(\vec{F}, \vec{T}),
\end{equation}
where $\mathcal{R}$ is the set of design rules, $\vec{T}$ is the topology and $\vec{F}$ is the physical size of layout pattern. When the legalization fails, the unreasonable region can be located and returned due to the explainable feature of the legalization function.

%% file: doc/result.tex
\section{Experimental Results}

\begin{table*}
\caption{
    Comparison on Legality and Diversity on legal patterns.
    %Results of previous works are from \cite{wang2023diffpattern} except \cite{wang2023diffpattern}.
    `/' refers to not applicable. 
}
\resizebox{0.92\textwidth}{!}{
\begin{threeparttable}[tb!]
    \centering
    
    \label{tab:diversity}
    
        \begin{tabular}{|c|c|c|c|cc|cc|cc|}
            \hline
            \multirow{2}*{Task}&\multirow{2}*{Set/Method}&\multirow{2}*{Training Set$^*$}&\multirow{2}*{Size}&\multicolumn{2}{c|}{Layer-10001}&\multicolumn{2}{c}{Layer-10003}&\multicolumn{2}{|c|}{Total$^\dagger$}\\
            &&&&Legality ($\uparrow$)& Diversity ($\uparrow$)  &Legality ($\uparrow$)& Diversity ($\uparrow$)&Legality ($\uparrow$)& Diversity ($\uparrow$)\\ 
            \hline
            \hline
            \multirow{6}{*}{\rotatebox{90}{Fixed-size}}& Real Patterns&/&\multirow{6}*{$128^2$}&/&10.731&/&8.769&/&10.625\\
            &CAE+LegalGAN~\shortcite{zhang2020layout}&Layer-10001&&3.74\%&5.814&/&/&/&/ \\
            &VCAE+LegalGAN~\cite{zhang2020layout}&Layer-10001&&84.51\%&9.867&/&/&/&/ \\
            &LayouTransformer~\shortcite{wen2022layoutransformer}&Layer-10001& &89.73\%&10.527&/&/&/&/ \\
            &DiffPattern~\shortcite{wang2023diffpattern}&Layer-10001/10003&&\textbf{99.97\%}&10.711&99.98\%&8.578&\textbf{99.98\%}&10.633\\

            &ChatPattern&Layer-10001/10003&&\textbf{99.97}\%&\textbf{10.796}&\textbf{99.99\%}&\textbf{8.625}&\textbf{99.98\%}&\textbf{10.650}\\
            \hline
            \hline
            \multirow{9}{*}{\rotatebox{90}{Free-size}}&Real Patterns&/&\multirow{3}*{$256^2$} & /&12.702&/&10.696&/&12.695 \\
            &\shortcite{wang2023diffpattern} w/ Concatenation &Layer-10001/10003& &57.78\% &10.719&93.69\%&10.511&75.74\%&11.706\\
            &\tool{ChatPattern}&Layer-10001/10003& &\textbf{87.36\%}&\textbf{11.154}&\textbf{99.78\%}&\textbf{10.556}&\textbf{93.57\%}&\textbf{11.830}\\
            % In-Painting&$256^2$&79.17\%&10.816&96.96\%&\textbf{10.736}&88.07\%&11.763\\
            % Out-Painting&$256^2$ &\textbf{87.36\%}&\textbf{11.154}&\textbf{99.78\%}&10.556&\textbf{93.57\%}&\textbf{11.830}\\
            \cline{2-10}
            &Real Patterns& /&\multirow{3}*{$512^2$} &/&13.435 &/&12.139&/&13.787 \\
            &\shortcite{wang2023diffpattern} w/ Concatenation&Layer-10001/10003 &&0.29\%&5.714&40.83\% &11.555&20.56\%&11.359 \\
          
            &\tool{ChatPattern}&Layer-10001/10003&&\textbf{36.42\%}&\textbf{10.401}&\textbf{98.86\%}&\textbf{11.620}&\textbf{67.64\%}&\textbf{12.133}\\
            % In-Painting&$512^2$&32.90\%&\textbf{10.053}&85.86\%&\textbf{11.721}&59.38\%&12.110\\
            % Out-Painting&$512^2$&\textbf{36.42\%}&10.401&\textbf{98.86\%}&11.620&\textbf{67.64\%}&\textbf{12.133}\\
            \cline{2-10}
            &Real Patterns& /&\multirow{3}*{$1024^2$}&/&13.573&/&12.644&/&14.109 \\
            &\shortcite{wang2023diffpattern} w/ Concatenation &Layer-10001/10003& &0.00\%&0.000&0.64\%&6.926&0.32\%&6.926\\
      
            &\tool{ChatPattern}&Layer-10001/10003&&\textbf{1.19\%}&\textbf{6.438}&\textbf{94.96\%}&\textbf{11.981}&\textbf{47.80\%}&\textbf{11.992}\\
            % In-Painting&$1024^2$&\textbf{1.19\%}&\textbf{6.438}&56.00\%&11.676&28.60\%&11.714\\
            % Out-Painting&$1024^2$&0.64\%&4.938&\textbf{94.96\%}&\textbf{11.981}&\textbf{47.80\%}&\textbf{11.992}\\
            \hline
        \end{tabular}
    
    \begin{tablenotes}
    % \item[$\dagger$] We forbid pattern selection and pattern modification in legalization phase for a fair comparison. 
    \item[$*$] All training datasets are the 128$\times$128 version.
    \item[$\dagger$] We collected generated samples from both Layer-10001/10003 and evaluated them together.
    % \item[$\dagger$] We re-implemented DiffPattern according to the paper~\cite{wang2023diffpattern} and subsequently trained two distinct models, each on a separate dataset.
    \end{tablenotes}
\end{threeparttable}
}
\end{table*}

\subsection{Layout Pattern Generation}

\minisection{Datasets.} We follow previous works~\cite{wang2023diffpattern,wen2022layoutransformer} to obtain the dataset of small layout pattern images with the size of 2048$\times$2048 $nm^2$ by splitting the layout map from ICCAD contest 2014. The size of the extracted topology matrix is fixed as 128$\times$128 in squish pattern representation in the baseline setting. There are two different style layouts, denoted as Layer-10001 and Layer-10003. Layer-10001 is widely used in previous methods and we introduce layer-10003 to evaluate the model ability of style specification. Furthermore, by splitting the layout map into 4$\times$, 16$\times$ and 64$\times$ larger size with overlap, we also obtained 4$\times$, 16$\times$ and 64$\times$ larger topology matrix in squish pattern representation correspondingly.

\begin{figure}[tb!]
    \vspace{-.2in}
    \centering
    \subfloat[Layer-10001 style]{\includegraphics[width=0.46\linewidth]{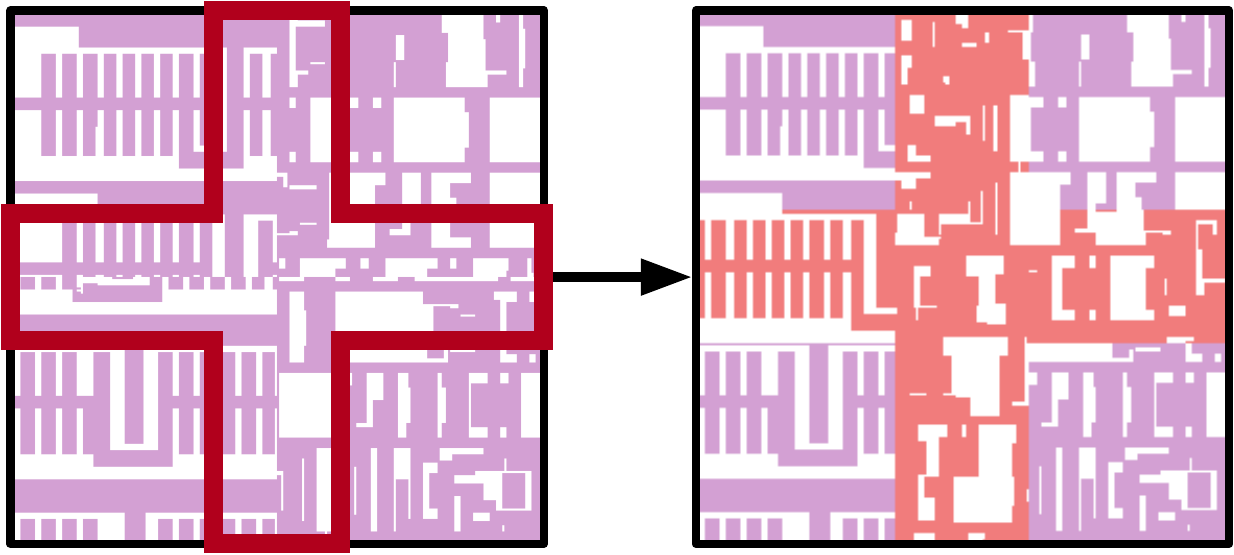}}
    \hfill
    \subfloat[Layer-10003 style]{\includegraphics[width=0.46\linewidth]{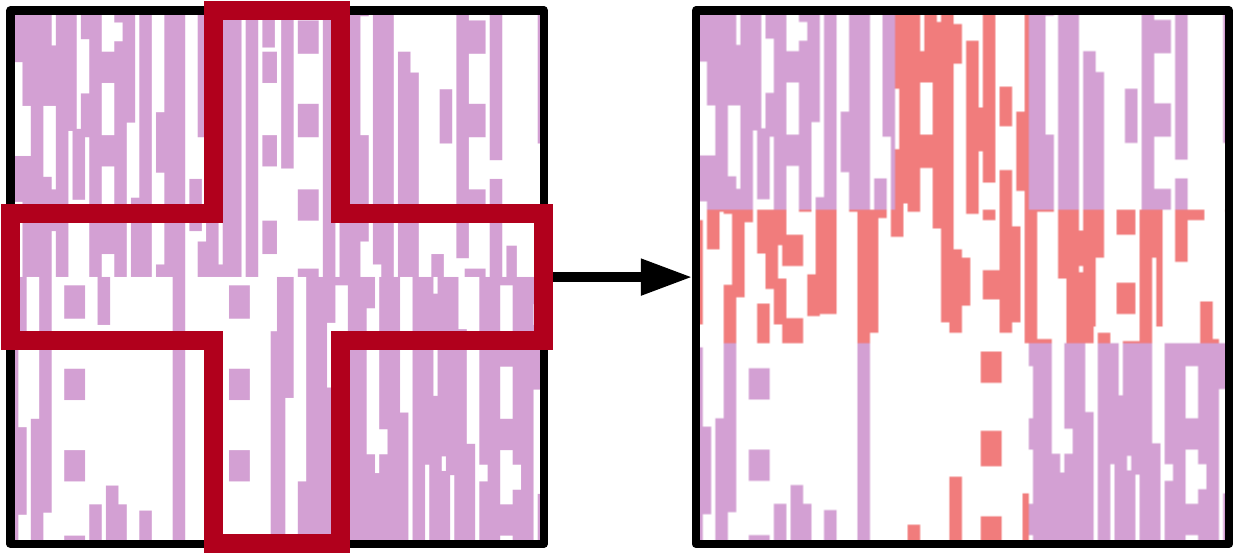}}
    \caption{Example of 256$\times$256 topology matrix generated by In-Painting (without legalization).}
    \label{fig:modification}
\end{figure}

\minisection{Diffusion Model Configuration.} Following the common settings~\cite{ho2020denoising,austin2021structured}, we use a U-Net~\cite{ronneberger2015u} as our backbone in the conditional discrete diffusion model. The implementation of U-net follows that in~\cite{ho2020denoising}.
To extract the embedding of the condition, we use a stack of 3 linear layers. The condition embedding is added into the embedding of the time step to control the style of the generated patterns.

\begin{figure}[tb!]
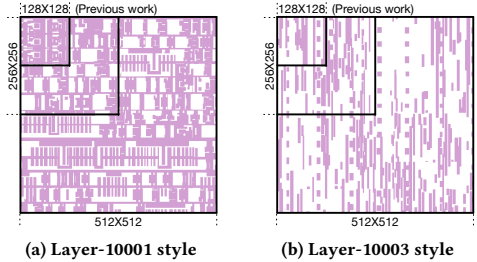

    \centering
    \subfloat[Layer-10001 style]{\includegraphics[width=0.328\linewidth]{outpaint-part1.pdf}}
    \hspace{.2in}
    \subfloat[Layer-10003 style]{\includegraphics[width=0.328\linewidth]{outpaint-part2.pdf}}
    \caption{512$\times$512 topology matrix generated by Out-Painting.}
    \label{fig:extension}
\end{figure}

\minisection{Training \& Testing Details.} To keep the same size with previous squish-pattern-based methods~\cite{wang2023diffpattern,zhang2020layout}, we train \tool{ChatPattern} on the 128$\times$128 union-datasets (Layer-10001 and Layer-10003) with class conditions for $1.0$M iterations with a batch size of $128$. The network is optimized by an Adam optimizer with a learning rate 2e-4. Drop out is $0.1$ the grad clip is set to $1$, and the loss coefficient $\lambda$ is set to 1e-3. Diffusion length $K=1000$ and the noise schedule $\beta_k$ is linearly increased from $0.01$ to $0.5$. The training procedure takes about $250$ GPU hours in total.
The back-end of \tool{ChatPattern} is trained on 128$\times$128 dataset and the size of directly generated typologies is fixed at 128$\times$128. The topology matrix sampled from the network will be further legalized as explained in \Cref{sec:3.3}. We have noticed previous work~\cite{wang2023diffpattern} applies topology selection and pushes the legality to 100\%. Since every squish-pattern-based method can reach 100\% legality via selection, we remove the selection step from all methods when evaluating their performance to compare the model directly. We further forbid our pattern modification function calling in both fixed-size and free-size pattern generation tasks to have a fair comparison.

\minisection{Fixed-size Pattern Generation.} We compare our method with previous layout pattern generation methods, CAE \cite{yang2019deepattern}, VCAE \cite{zhang2020layout}, LegalGAN \cite{zhang2020layout} and LayouTransformer \cite{wen2022layoutransformer} on the widely used benchmark 128$\times$128 Layer-10001. And we further re-implemented the previous SOTA, DiffPattern\cite{wang2023diffpattern}, on Layer-10001 and Layer-10003 individually since directly mixing patterns from different distributions can easily lead to a conflict for \cite{wang2023diffpattern}, as we discussed in \Cref{sec:3.3}. We report the result on 10,000 samples generated by \tool{ChatPattern} for each class. The legality and the diversity of legal patterns on each dataset are reported in \Cref{tab:diversity}. According to the results, on the 128$\times$128 pattern generation task, \tool{ChatPattern} gets a reasonable improvement compared with previous SOTA thanks to its ability to utilize a multi-source training dataset. However, both \tool{ChatPattern} and the existing models have already fitted the distribution of real patterns well in the simple 128$\times$128 benchmarks.

\begin{figure}[tb!]
    \centering
   \includegraphics[width=0.88\linewidth]{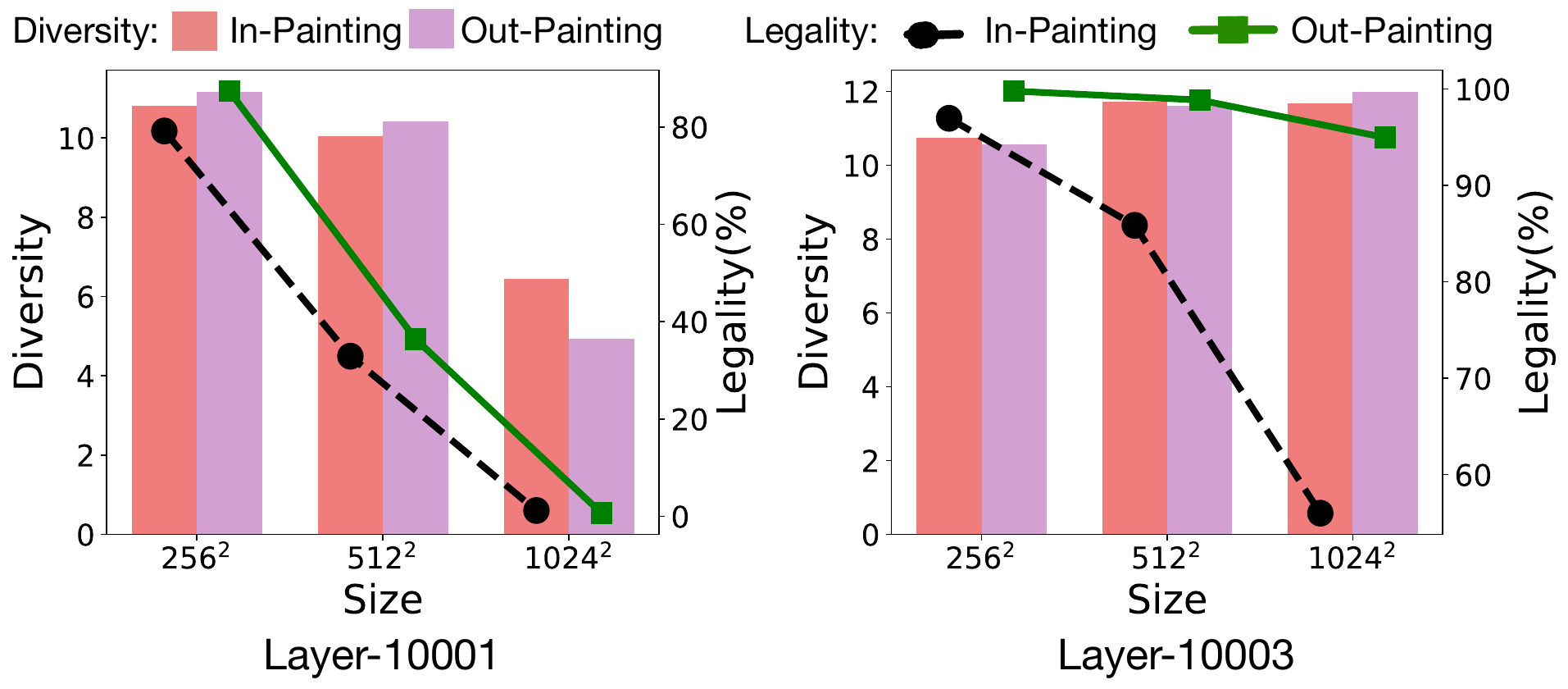}
    \caption{Evaluation of In-Painting and Out-Painting.}
    \label{fig:abla}
\end{figure}

\minisection{Free-size Pattern Generation.} \tool{ChatPattern} is specialized in free-size pattern generation, a task highly applicable to real-world scenarios. For this task, we established three distinct experimental size settings to quantitatively assess the model. The sizes of the target topology matrices ranged from $256^2$ to $1024^2$. We generated 10,000 samples for each class across all size levels, examining their Legality and Diversity as we did in the fixed-size pattern generation tasks. Additionally, patterns extracted from real datasets served as references and are duly noted in the results table.
The baseline is the SOTA in fixed-size pattern generation, DiffPattern~\cite{wang2023diffpattern}. To create larger patterns, the baseline method can only stitches together small patches of the same class, but the method often breaches design rules. For instance, when patch size reaches $512^2$, the legality of patterns generated by DiffPattern with concatenation plummets to nearly zero (0.29\%) and under half (40.83\%) for the Layer-10001 and Layer-10003 datasets, respectively.
Our tests with \tool{ChatPattern} cover all the aforementioned settings. \tool{ChatPattern} leverages its extension function, utilizing both out-painting and in-painting algorithms, to synthesize high-quality patterns. We have illustrated some instances in \Cref{fig:modification} and \Cref{fig:extension}. The LLM agent is initialized with a predefined working pipeline, as depicted in \Cref{fig:llm}. The accompanying documentation includes statistical data on pattern extension through various algorithms, as seen in \Cref{fig:abla}. This documentation provides an insight that out-painting typically yields better legality, while in-painting excels in diversity under certain conditions. The user's request was simply to generate 10,000 patterns for each experimental setting. Results tabulated in \Cref{tab:diversity} demonstrate \tool{ChatPattern}'s superiority over the baseline, particularly in the demanding task of generating large patterns of sizes $512^2$ and above.

\subsection{Evaluation of LLM agent}

We take the user requirement in \Cref{fig:llm} as a running example to show the LLM agent's ability on requirement auto-formatting and unseen mistake-processing.

\minisection{Requirement Auto-formatting.}
The LLM agent is required to fill out a standard requirement list to make sure the target of every sub-task is specific and can be handled within a simpler script. An example of the requirement list is following. %By splitting the task into sub-tasks before task-planing, every sub-task can be handled within a simpler script.
\begin{mdframed}
    \small
\# Requirement - subtask 1

\noindent\#\# Basic Part: 
Topology Size: [200, 200], 
Physical Size: [1500, 1500] nm, 
Style: Layer-10001, 
Count: 50000, 

\noindent\#\# Advanced Part: 
Extension Method: Out (Default: Out), 
Drop Allowed: True (Default: True), 
Time Limitation: None (Default: None).
\end{mdframed}

\minisection{Unseen Mistake-processing.}
As explored in \Cref{Sec:algo}, the generation of extensive topology matrices poses significant time demands. Should the legalization process fail, the function earmarks the errors within a log. We present a case where the LLM agent adeptly navigates unanticipated mistakes during legalization. It is noteworthy that such error-handling expertise is not pre-coded within the standard operational procedures or the experiential documentation. This instance serves to highlight the LLM agent's intrinsic capacity to address unforeseen complications.

\begin{mdframed}
    \small
\% In previous attempts, one pattern can't pass the legalization.

\noindent \textbf{Thought}: Since legalization has failed twice in the same region, I will try to in-paint that specific area with the same style and then attempt legalization again.

\noindent \textbf{Action}: Topology\_Modification

\noindent \textbf{Action Input}: {
"topology\_path":\$\{path\},
"upper": 12,
"left": 56,
"bottom": 33,
"right": 73,
"style": "Layer-10001",
"seed": 42

\noindent\% Continue Processing
}

\end{mdframed}

%% file: doc/conclusion.tex
\section{Conclusion}

In this paper, we introduced \tool{ChatPattern}, a novel framework for pattern generation utilizing a LLM. \tool{ChatPattern} provides a user-friendly interface that accepts natural language inputs to tailor the pattern library to specific needs. And \tool{ChatPattern} enables a flexible pattern generation back-end and extends the task of pattern generation to a more demanding yet challenging setting, free-size pattern generation. Nonetheless, \tool{ChatPattern} still lacks global guidance when generating large patterns and can not handle complex oversized patterns well, which we left for future exploration.